\documentclass[letterpaper, 10 pt, conference]{ieeeconf}  
\IEEEoverridecommandlockouts
\pdfoutput=1
\usepackage{threeparttable}
\usepackage{cite}
\usepackage{amsmath,amssymb,amsfonts}
\usepackage{graphicx}
\usepackage{textcomp}
\usepackage{color}
\usepackage{xcolor}
\usepackage{colortbl}
\usepackage{subfigure}
\usepackage{multirow}
\usepackage{multicol}
\usepackage{makecell}
\usepackage{bm}
\usepackage{algorithm}
\usepackage{algpseudocode}
\usepackage{booktabs} 
\usepackage[left=48pt, right=48pt, top=57pt, bottom=43pt]{geometry}
\usepackage[colorlinks,bookmarks,bookmarksopen,bookmarksnumbered,linktocpage,hyperindex,citecolor=blue, linkcolor=blue, urlcolor=blue]{hyperref}

\newcommand{\Reffig}[1]{Fig.~\ref{#1}}
\newcommand{\Refsec}[1]{Section~\ref{#1}}

\newcommand{\Reftab}[1]{Table~\ref{#1}}

\begin{document}
\title{
        Ranking-aware Continual Learning for LiDAR Place Recognition
}

\author{Authors}
\author{Xufei Wang$^{1}$, Gengxuan Tian$^{2}$, Junqiao Zhao$^{*, 1, 2, 3}$, Siyue Tao$^{2}$, Qiwen Gu$^{2}$, Qiankun Yu$^{4}$, Tiantian Feng$^{5}$ 
\thanks{This work is supported by XXX. \emph{(Corresponding Author: Junqiao Zhao.)}}
\thanks{$^{1}$Xufei Wang, Junqiao Zhao are with the Shanghai Research Institute for Intelligent Autonomous System, Tongji University, Shanghai, China
{\tt\footnotesize (Corresponding Author: zhaojunqiao@tongji.edu.cn).}}
\thanks{$^{2}$Gengxuan Tian, Siyue Tao, Qiwen Gu and Junqiao Zhao are with the Department of Computer Science and Technology, School of Electronics and Information Engineering, Tongji University, Shanghai, China, and the MOE Key Lab of Embedded System and Service Computing, Tongji University, Shanghai, China}
\thanks{$^{3}$Institute of Intelligent Vehicles, Tongji University, Shanghai, China}
\thanks{$^{4}$SAIC Intelligent Technology (Shanghai) Co. Ltd, Shanghai, China}
\thanks{$^{5}$College of Surveying and Geo-Informatics, Tongji University, Shanghai, China}
}

\maketitle

\begin{abstract}
Place recognition plays a significant role in SLAM, robot navigation, and autonomous driving applications.
Benefiting from deep learning, the performance of LiDAR place recognition (LPR) has been greatly improved.
However, many existing learning-based LPR methods suffer from catastrophic forgetting, which severely harms the performance of LPR on previously trained places after training on a new environment. 
In this paper, we introduce a continual learning framework for LPR via Knowledge Distillation and Fusion (KDF) to alleviate forgetting. 
Inspired by the ranking process of place recognition retrieval, we present a ranking-aware knowledge distillation loss that encourages the network to preserve the high-level place recognition knowledge.
We also introduce a knowledge fusion module to integrate the knowledge of old and new models for LiDAR place recognition.
Our extensive experiments demonstrate that KDF can be applied to different networks to overcome catastrophic forgetting, surpassing the state-of-the-art methods in terms of mean Recall@1 and forgetting score.

\end{abstract}

\section{INTRODUCTION}
\label{sec:introduction} 
\begin{figure}[t]
        \centering
        \includegraphics[width=\columnwidth]{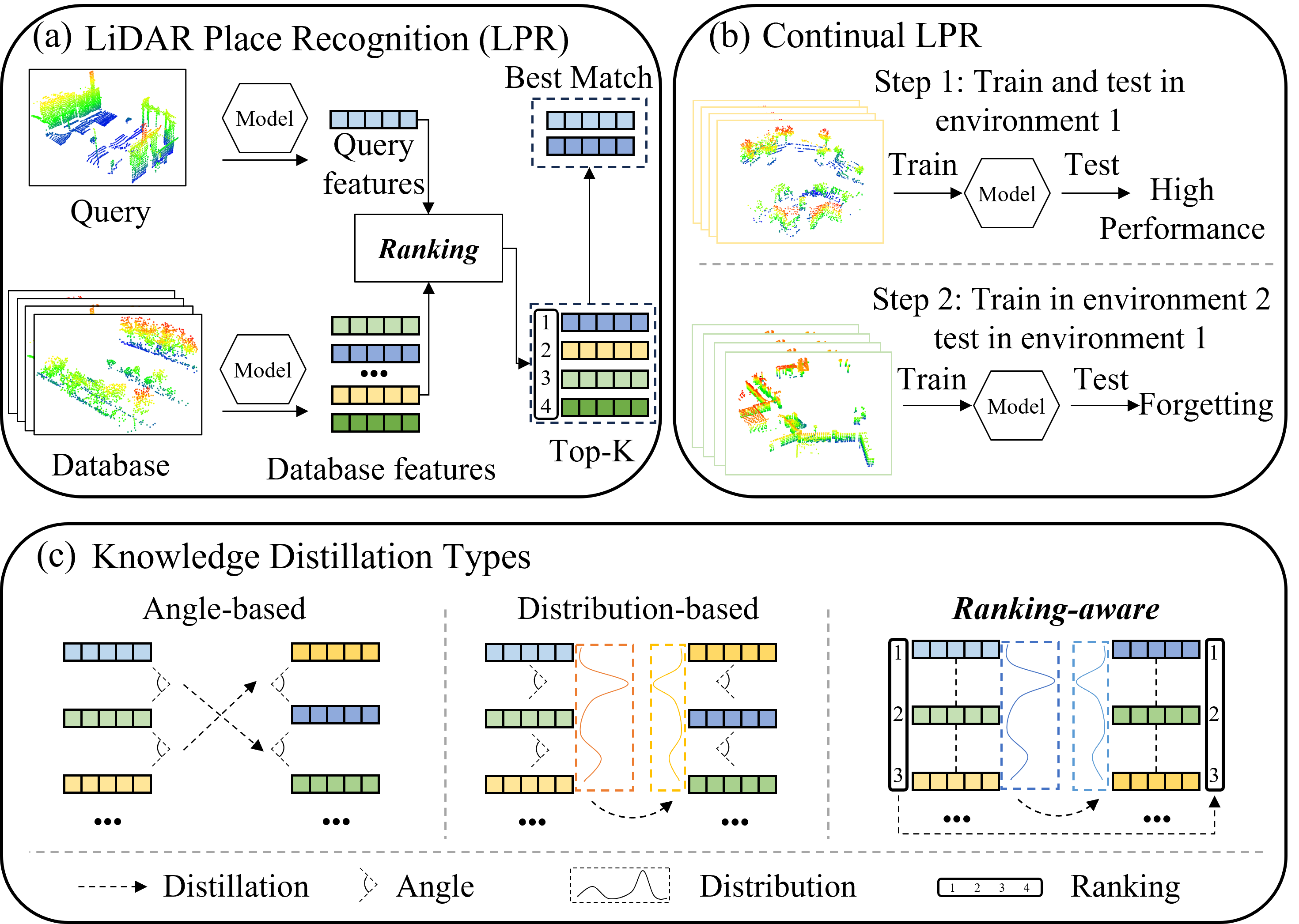}
        \caption{The LiDAR place recognition (LPR) and continual LPR, and comparison of knowledge distillation strategies in different methods. (a) The LiDAR place recognition (LPR) get the best match of query sample by the ranking-based retrieval. (b) The LPR model learns from different environments and the catastrophic forgetting lead to performance degradation on old environment. (c) Comparison of knowledge distillation strategies between existing continual learning methods for LPR and our proposed method. Existing methods rely solely on response alignment between old and new models, such as angles (InCloud \cite{knights2022irosincloud}) or distributions (CCL \cite{cui2023ralccl}) between descriptors. In contrast, our method not only models the feature distributions of the old and new models but also captures the intrinsic relationship between features by incorporating ranking consistency.
        }\label{fig:continual LPR}
        \vspace{-0.5cm}
\end{figure}

Place recognition (PR) aims to determine a robot's location by matching its current observations with a database of previously visited places. 
It plays a significant role in SLAM, robot navigation, and autonomous driving applications \cite{yin2024survey, yin2024general, shi2023lidar}.
Leveraging the strong representation learning capabilities of deep learning, high-performance learning-based LiDAR place recognition (LPR) methods have demonstrated robustness to appearance variations in places \cite{uy2018pointnetvlad, komorowski2021minkloc3d, vidanapathirana2022logg3d, komorowski2022improving, liu2019lpd, chen2022overlapnet, xu2023transloc3d} (\Reffig{fig:continual LPR} (a)). 

Recently, it has been acknowledged that PR systems must continuously learn from diverse environments \cite{yin2023bioslam}. 
When transitioning between cities, the PR system's capabilities should be incrementally updated. 
However, as new data continuously streams in, retraining the model becomes computationally expensive. 
Furthermore, neural networks are susceptible to overfitting to the data distribution of new environments, resulting in catastrophic forgetting \cite{wang2024comprehensive}, i.e., performance degradation on previously learned environments, as illustrated in \Reffig{fig:continual LPR} (b).

To address these challenges, recent studies \cite{knights2022irosincloud, cui2023ralccl} have introduced continual learning to maintain the performance of LPR systems while minimizing storage requirements. 
To mitigate catastrophic forgetting, InCloud \cite{knights2022irosincloud} employs structure-aware (angle-based) knowledge distillation to preserve the embedding structures of different environments. 
Similarly, CCL \cite{cui2023ralccl} constructs a contrastive learning feature pool and applies knowledge distillation to maintain consistent feature distributions across environments. 

Despite their successes, these methods primarily focus on response alignment between old and new models, such as angles or distributions between descriptors (\Reffig{fig:continual LPR} (c)). 
However, they overlook the intrinsic relationship alignment of descriptors between the old and new models. 
Moreover, the old model is discarded once knowledge distillation is completed, without leveraging the hidden knowledge it may contain. 

Building on the observations above, we propose a novel continual learning framework for LPR via Knowledge Distillation and Fusion, termed KDF. 
Inspired by the retrieval strategy in LPR methods \cite{uy2018pointnetvlad, komorowski2021minkloc3d, vidanapathirana2022logg3d, komorowski2022improving, liu2019lpd, chen2022overlapnet}, where previously visited places are ranked based on the similarity between a query and the database, we incorporate ranking information as an intrinsic relationship for knowledge distillation. 
This leads to the development of a ranking-aware knowledge distillation loss function, which not only measures overall distributional differences between old and new features but also captures their ranking discrepancies. 
Furthermore, to fully exploit the rich knowledge embedded in the old model, we introduce a knowledge fusion module. 
This module integrates knowledge from both the old and new place recognition models, significantly mitigating model forgetting. 

To summarize, the main contributions of this paper are as follows:
\begin{itemize}
        \item We present a ranking-aware knowledge distillation loss, which encourages the network to preserve the high-level place recognition knowledge.
        \item We also introduce a knowledge fusion module to integrate the knowledge of old and new models for LiDAR place recognition.
        \item Through extensive experiments, we demonstrate the superior performance of KDF, which can be integrated into various networks to overcome catastrophic forgetting.
\end{itemize}

\section{RELATED WORKS}
\label{sec:related works}
\subsection{LiDAR Place Recognition}

Over the past decade, place recognition research has experienced significant growth \cite{yin2024survey, shi2023lidar, yin2024general}. 
Existing LPR methods can be broadly categorized into handcrafted and learning-based approaches. 

Handcrafted LPR methods typically design descriptors based on attributes such as point cloud height and intensity \cite{rohling2015fast, kim2018scan, wang2020intensity}. 
\cite{rohling2015fast} constructs a one-dimensional histogram descriptor by directly counting the height distribution of the point cloud. 
Scan Context (SC) \cite{kim2018scan} projects 3D point clouds onto a 2D grid, where each bin's value is determined by the maximum height within its segment. 
Building on SC, \cite{wang2020intensity} incorporates intensity information into descriptor construction and introduces a two-stage hierarchical intensity retrieval strategy. 
These methods are computationally efficient and easy to implement. However, their performance and robustness heavily depend on parameter tuning. 

Compared to handcrafted methods, learning-based LPR approaches can extract more representative and robust scene descriptors. 
PointNetVLAD \cite{uy2018pointnetvlad} employs PointNet \cite{qi2017pointnet} to extract local point cloud features, which are then aggregated into global descriptors using NetVLAD \cite{arandjelovic2016netvlad}. 
MinkLoc3D \cite{komorowski2021minkloc3d} leverages a 3D feature pyramid architecture based on sparse voxels and sparse 3D convolutions to extract local features, followed by a GeM \cite{radenovic2018fine} pooling layer to generate global descriptors. 
MinkLoc3Dv2 \cite{komorowski2022improving} enhances this framework by incorporating additional convolution and transposed convolution blocks, increasing network depth and width through a higher number of channels. 
It also refines the loss function using a differentiable average precision approximation \cite{brown2020smooth} combined with multi-level backpropagation. 
LoGG3D-Net \cite{vidanapathirana2022logg3d} employs a sparse U-Net to embed each point into a high-dimensional feature space and introduces a local consistency loss to maximize feature similarity. 
It further utilizes second-order pooling with differentiable eigenvalue power normalization to obtain a global descriptor. 
TransLoc3D \cite{xu2023transloc3d} introduces an adaptive receptive field module with point-wise feature reweighting. 
It employs a transformer module to capture long-range feature dependencies and utilizes a NetVLAD \cite{arandjelovic2016netvlad} layer to generate global descriptors. 

While these methods achieve state-of-the-art performance, they suffer from catastrophic forgetting when trained on new environments, leading to a decline in performance on previously learned environments. 

\subsection{Continual Learning}
Continual learning \cite{wang2024comprehensive, shaheen2022continual, lesort2020continual}, also referred to as incremental learning or lifelong learning, has been introduced to address the problem of catastrophic forgetting. It can be broadly categorized into three main approaches. 

The first group consists of replay-based or rehearsal-based methods, which select a set of old domain samples \cite{rebuffi2017icarl} or pseudo-memory samples \cite{shin2017continual} generated by the learning model and store them in a replay buffer. 
During training, replaying the samples from the buffer enables the model to retain knowledge from previous domains, thereby alleviating the forgetting problem. 

The second group consists of regularization-based methods, which introduce explicit regularization terms to balance the learning of old and new tasks. Within this group, weight regularization methods, such as EWC \cite{kirkpatrick2017overcoming} and SI \cite{zenke2017continual}, selectively regularize the variation of network parameters. 
Function regularization methods, on the other hand, target the intermediate features or final output of the prediction function. 
These methods typically use the previously learned model as the teacher and the current model as the student, applying knowledge distillation (KD) \cite{gou2021knowledge, park2019relational} to mitigate catastrophic forgetting. 

The third group consists of architecture-based methods \cite{serra2018overcoming, mallya2018packnet}, which dynamically modify the model's architecture by freezing or isolating task-specific parameters. 
The new task is learned by adding new modules to the model, while the previously learned parameters remain unchanged.

Although continual learning has been extensively explored in fields such as image classification \cite{li2017learning, shin2017continual, rebuffi2017icarl, mallya2018packnet}, segmentation \cite{shi2022incremental, stan2021unsupervised}, and object detection \cite{zhao2022static, wang2021wanderlust}, only a few studies \cite{gao2022airloop, yin2023bioslam, knights2022irosincloud, cui2023ralccl, liu2024micl} have attempted to apply continual learning to place recognition. 

Airloop \cite{gao2022airloop} is the first approach to consider lifelong learning for visual loop closure detection, using a regularization-based strategy with Euclidean-distance knowledge distillation. 
BioSLAM \cite{yin2023bioslam} proposes a lifelong learning place recognition (PR) framework, which includes a gated generative replay mechanism with dynamic and static memory zones. 
However, these methods address the forgetting problem under varying lighting, weather, and seasonal conditions, without considering continual learning across different environments. 

InCloud \cite{knights2022irosincloud} integrates continual learning into LPR for the first time, employing an angular distillation loss to preserve the structure of the embedding space across different environments. 
CCL \cite{cui2023ralccl} utilizes asymmetric contrastive learning for LPR and introduces a feature distribution-based knowledge distillation loss for past samples to train more transferable place representations. 
However, these methods simply apply similar distillation losses \cite{park2019relational} from other domains to continual LPR. 
Recently, MICL \cite{liu2024micl} introduces mutual information into continual learning for LPR, aiming to preserve more domain-shared information through a proposed mutual information loss.

In contrast to these approaches, we propose a novel ranking-aware knowledge distillation loss to specifically transfer knowledge for LPR. 
Additionally, we integrate knowledge between the old and new models by introducing a knowledge fusion module.

\begin{figure*}[t!]
        \centering
        \includegraphics[width=.95\textwidth]{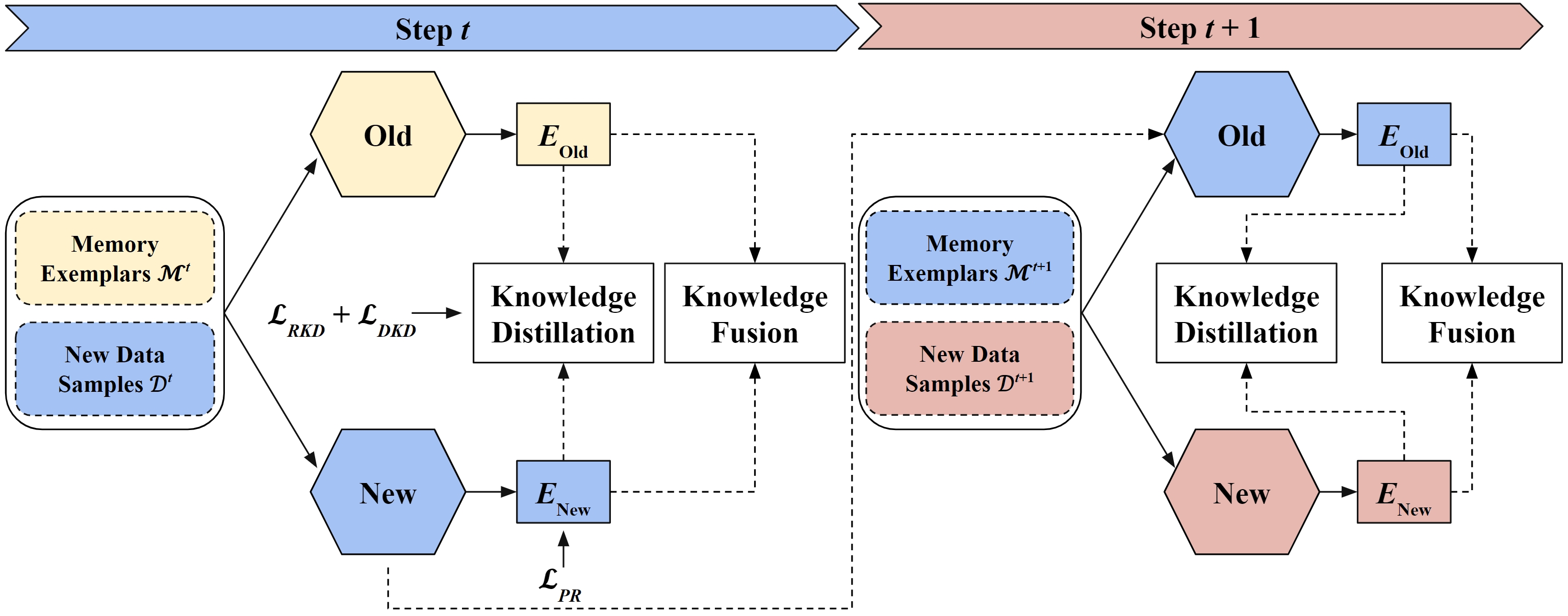}
        \caption{An overview of our proposed method KDF.
        We first initialize a new model from the old model and freeze the old model. 
        Next, memory exemplars and new data are fed into both models for knowledge distillation and metric learning. 
        After training, the acquired knowledge from both models is fused in preparation for future steps. 
        Let $E_{old}$ and $E_{new}$ denote the embeddings from the old and new models, respectively. 
        $\mathcal{L}_{PR}$ represents the place recognition loss, i.e., the triplet margin loss. 
        $\mathcal{L}_{RKD}$ and $\mathcal{L}_{DKD}$ correspond to the ranking-based distillation loss and the distribution-based distillation loss, respectively. 
        Yellow, blue, and red indicate steps $t-1$, $t$, and $t+1$, respectively.
        }\label{fig:pipeline}
        \vspace{-0.5cm}
\end{figure*}

\section{PRELIMINARIES}
\label{sec:problem formulation}
Before introducing our method, we first define the LPR task and then formulate the problem of continual learning for LPR.

\subsection{LiDAR Place Recognition}
Given a query point cloud $X$ and a reference database $D = \{Y_1, Y_2, \dots, Y_n\}$, the LPR task can be formulated as:
\begin{equation}
        \hat{Y} = \mathop{\arg\max}\limits_{Y} score (X,Y\vert{Y \in{D}} ) ,
\end{equation}
where $Y$ represents a matching candidate for $X$ in the database $D$, and the $score$ is the similarity function.

Most LPR methods train a model parameterized by $\varTheta$ that encodes the query point cloud $X$ and the reference database $D$ into a global feature descriptor $f_X$ and a set of database feature descriptors $\{f_{Y_1},f_{Y_2},\dots,f_{Y_n}\}$.
The similarity or matching score is then obtained by calculating the distance between $f_X$ and $f_{Y_i}$. 
Ultimately, the location of the best matching point cloud scan is used as the location of the query point cloud $X$.

\subsection{Continual LiDAR Place Recognition}
In the continual LPR setting, we aim to equip the LPR model with the ability to learn generalizable knowledge from $N$ sequentially incoming LPR datasets, $\mathcal{D} = \{\mathcal{D}^i\}_{i=1}^N$.
The datasets may be spatially disjoint, such that each dataset represents a unique domain.
During each training step $t$, the network is initialized using the training parameters $\varTheta_{t-1}$ from the previous step and is trained on $\mathcal{D}^t \cup \mathcal{M}^t$, where $\mathcal{M}^t$ denotes the memory buffer, which stores a limited number of exemplars from each previous domain.

\section{METHODOLOGY}
\label{sec:methodology}
Pioneering works such as InCloud \cite{knights2022irosincloud} and CCL \cite{cui2023ralccl} introduce continual learning approaches for LiDAR place recognition (LPR), explicitly combining memory replay strategies with knowledge distillation. 
However, these methods do not account for the essential ranking information within the learned representations.

To address this limitation, we propose a novel continual learning framework with ranking-aware distillation for LPR.
As illustrated in \Reffig{fig:pipeline}, our method consists of two key components: Knowledge Distillation and Knowledge Fusion.
To efficiently transfer knowledge from the old model to the new one, we introduce a ranking-aware knowledge distillation loss, which minimizes discrepancies in the embedding space while preserving ranking information.
Additionally, to further mitigate catastrophic forgetting and enhance the model's generalization ability, we integrate old and new knowledge within the embedding space through knowledge fusion.


\subsection{Ranking-Aware Knowledge Distillation}
Our ranking-aware knowledge distillation is composed of two components: the ranking-based knowledge distillation and the distribution-based knowledge distillation. 

\subsubsection{Ranking-based Knowledge Distillation}

\begin{figure}[t]
        \centering
        \includegraphics[width=.95\columnwidth]{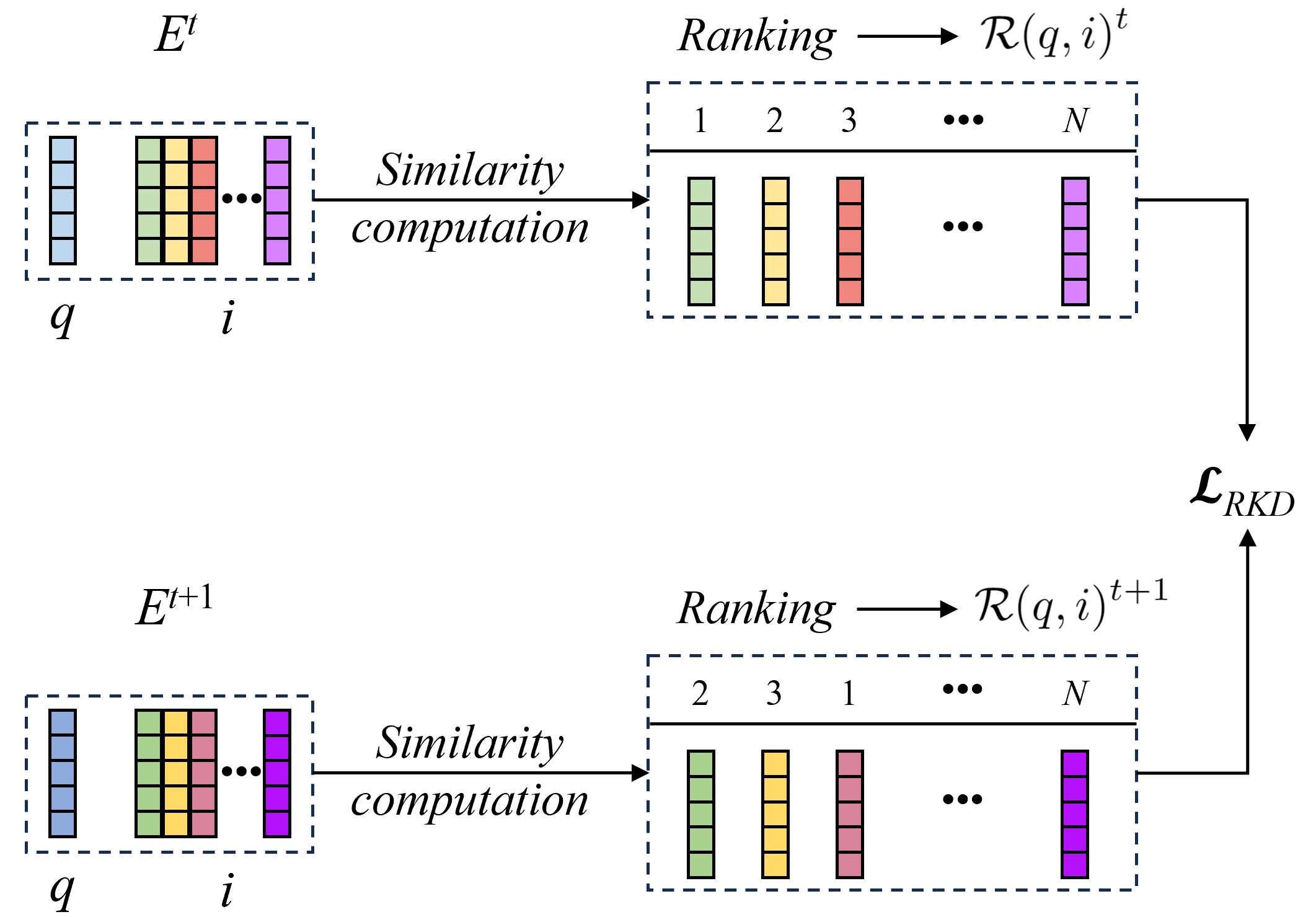}
        \caption{The ranking-based knowledge distillation loss $\mathcal{L}_{RKD}$ enforces the model to maintain more informative knowledge from the previous domain. Given two embedding sets $E^{t}$ and $E^{t+1}$ from the old and new models, we compute their internal ranking information $\mathcal{R}(q,i)^{t}$ and $\mathcal{R}(q,i)^{t+1}$ by similarities. Then, we minimize the difference between them by knowledge distillation. 
        }\label{fig:ranking loss}
        \vspace{-0.5cm}
\end{figure}

We propose a novel ranking-based knowledge distillation loss (as illustrated in \Reffig{fig:ranking loss}), which enforces consistency in the ranking structure of embeddings generated by both the old and new models. 
However, the original ranking information is discrete, making it unsuitable for direct optimization.

Inspired by the Smooth AP loss \cite{brown2020smooth} used in Minkloc3Dv2 \cite{komorowski2022improving}, we incorporate a ranking function to address this challenge, ensuring that the ranking-based knowledge distillation loss remains compatible with standard optimizers.

Given a number of $N_b$ mini-batch samples and their embedding sets $E$, the ranking of sample $i$ corresponding to a query sample $q$ is defined as:
\begin{equation}
        \mathcal{R}(q,i) = 1 + \sum_{j\in N_b,j\neq i} \mathcal{G} (D(i,j);\tau ),
        \label{eq:Soft_ranking_list}
\end{equation}
where $D(i,j)$ is a difference matrix, defined as
\begin{equation}
        D(i,j) = S(q,j) - S(q,i),
\end{equation}
where $S \in \mathbb{R}^{N_b\times N_b}$ is the pairwise similarities' matrix of the embedding sets $E$.
In line with Minkloc3Dv2 \cite{komorowski2022improving}, we use Euclidean distance to calculate the similarity between features. 
$\mathcal{G} (\cdot;\tau )$ is a sigmoid function, defined as \begin{equation}
        \mathcal{G} (x;\tau ) = \frac{1}{1+e^{\frac{-x}{\tau} }},
\end{equation}
and $\tau$ is the temperature parameter. 

Finally, we compute the internal ranking information $\mathcal{R}(q,i)^{t}$ and $\mathcal{R}(q,i)^{t+1}$ according to embedding sets $E^{t}$ and $E^{t+1}$ and minimize the difference between them. 
This ranking-based knowledge distillation loss is formulated as:
\begin{equation}
        \mathcal{L}^t_{RKD}  = \frac{1}{N^3_b}\sum_{q = 1}^{N_b}\sum_{i = 1}^{N_b}\vert\mathcal{R}(q,i)^{t+1} - \mathcal{R}(q,i)^{t} \vert.
\end{equation}

\subsubsection{Distribution-based Knowledge Distillation}
In addition to ranking-based knowledge distillation, we further introduce feature distribution-based knowledge distillation for continual LPR.

Kullback-Leibler (KL) divergence \cite{kullback1951information} is a common metric in knowledge distillation, but it is an asymmetric divergence loss.
It measures the difference of distribution $P$ relative to $Q$, but not vice versa.
In knowledge distillation, this can lead to asymmetric information transfer.

Thus, we use a symmetric Kullback-Leibler (SKL) divergence loss to measure the difference in distributions, which is defined as follows:
\begin{equation}
        D_{SKL}(p\Vert q)  = \frac{1}{2} (D_{KL}(p\Vert q)+D_{KL}(q\Vert p)),
\end{equation}
where $D_{KL}$ is the KL Divergence. Thus, the distribution-based knowledge distillation loss is formulated as:
\begin{equation}
        \mathcal{L}^t_{DKD}  = \sum_{i = 1}^{N} D_{SKL} (E^t_i\Vert E^{t+1}_i),
\end{equation}
where $E^t$ and $E^{t+1}$ are the embeddings from the old and new models.
We will discuss the effectiveness of different divergences in \Refsec{sec:experiment_divergences}.

\subsection{Metric Learning for Place Recognition}

We use a triplet margin loss for training the LPR encoder, which is defined as follows:
\begin{equation}
        \mathcal{L}^t_{PR}  = \max(\Vert a^t,p^t\Vert_2 - \Vert a^t,n^t \Vert_2 + m,0),
\end{equation}
where $a^t, p^t, n^t$ are the embeddings at training step $t$ of a triplet of anchor sample, positive sample and negative sample from point clouds $\{P_a, P_p, P_n\}$, and $m$ is the margin parameter.

To achieve a better trade-off between stability and adaptability during the training process, we employ the distillation relaxation strategy proposed in InCloud \cite{knights2022irosincloud}.
Therefore, the overall loss of our method has the form of:
\begin{equation}
        \mathcal{L} = \mathcal{L}_{PR} + \lambda (\mathcal{L}_{RKD} + \mathcal{L}_{DKD}),
\end{equation}
where $\lambda$ is the decayed distillation relaxation parameter, defined as follows:
\begin{equation}
        \lambda = \frac{1}{1 + e^{10 \cdot   \frac{\gamma}{\beta  - 0.5} }},
\end{equation}
where $\gamma$ is the current training epoch, and $\beta$ is the total number of training epochs.
As training progresses, $\lambda$ gradually decreases, thereby reducing the contribution of the knowledge distillation loss to the overall objective.

\subsection{Knowledge Fusion}

Inspired by the dual-model mechanism in \cite{yu2023lifelong}, we introduce a knowledge fusion module into our framework.
The new model functions as the working model, while the old model serves as the memory model.
The working model processes new knowledge and absorbs information from the memory model.
Meanwhile, the memory model retains knowledge from previous environments and provides meta-knowledge about place recognition to the new model.
The knowledge fusion module integrates the knowledge from both models within the embedding space to fully leverage their place recognition capabilities.

Specifically, we obtain two features $f^t$ and $f^{t+1}$ for a query point cloud from the trained feature extractors $\varTheta^{t}$ and $\varTheta^{t+1}$. By concatenating these features, the fused representation is used for place recognition as follows: 
\begin{equation}
        f^{fuse} =  \{f^t \oplus  f^{t+1}\}.
\end{equation}

\section{EXPERIMENT}
\label{sec:experiment}
In this section, we first describe the datasets used in our experiments and the evaluation protocols.
We then compare the performance of our method to that of state-of-the-art continual learning methods.
Additionally, we present an ablation study to validate each component of KDF.
\subsection{Datasets and Experimental Protocol}
To evaluate the performance of KDF, we select four large point cloud datasets for our experiments: Oxford RobotCar \cite{maddern20171}, MulRan \cite{kim2020mulran}, In-house \cite{uy2018pointnetvlad}, and KITTI \cite{geiger2013vision}.
For the MulRan \cite{kim2020mulran} dataset, we choose two separate subsets, DCC and Riverside, for our continual LPR experiments.
\Reftab{tab:datasets} provides detailed information about the datasets and the training/testing configurations.
We preprocess all datasets following the procedures used in previous place recognition studies \cite{knights2022irosincloud, cui2023ralccl}.
Specifically, we remove the ground points from the point cloud scans and downsample the remaining points to 4096, normalizing the coordinates within the range [-1, 1].
We utilize two experimental protocols to analyze the performance of continual learning methods for LiDAR place recognition.
The first protocol is designed to evaluate the method's ability to overcome catastrophic forgetting. 
In line with the $4$-$Step$ protocol in InCloud \cite{knights2022irosincloud} and CCL \cite{cui2023ralccl}, we sequentially train on four datasets in the following order $Oxford\rightarrow DCC \rightarrow Riverside\rightarrow In$-$house$ and evaluate the method's performance step by step.
The second protocol is designed to evaluate the method's generalization capability. 
After continual training on four datasets, we directly evaluate the performance of different methods on sequences 00, 02, 07, and 08 from the KITTI dataset \cite{geiger2013vision}.

In line with the $4$-$Step$ protocol in InCloud \cite{knights2022irosincloud} and CCL \cite{cui2023ralccl}, we consider two scans as positive pairs if their distance is less than 10m, and as negative pairs if their distance exceeds 50m during training on the Oxford \cite{maddern20171} and In-house \cite{uy2018pointnetvlad} datasets.
During testing, pairs within a 25m distance are considered positive pairs.
For MulRan \cite{kim2020mulran}, the thresholds for positive and negative pairs during training are 10m and 20m, respectively.
In testing, positive pairs are defined as those within a 10m distance.
For KITTI testing, we regard a retrieved point cloud as a positive match if it is within 3m of the query.

We select four representative LPR network models to validate the performance of KDF: PointNetVLAD \cite{uy2018pointnetvlad}, LoGG3D-Net \cite{vidanapathirana2022logg3d}, MinkLoc3D \cite{komorowski2021minkloc3d}, and TransLoc3D \cite{xu2023transloc3d}.
Additionally, we compare KDF against FineTuning and five other continual learning methods.

\textbf{Fine-tuning}: The network is trained sequentially for each new environment without employing continual learning techniques.

\textbf{LwF} \cite{li2017learning}: A continual learning method that leverages knowledge distillation with KL divergence across training steps.

\textbf{EWC} \cite{kirkpatrick2017overcoming}: A regularization method that utilizes the Fisher Information Matrix to identify key parameters and penalizes their changes during training.

\textbf{InCloud} \cite{knights2022irosincloud}: A pioneering approach to continual LPR that integrates metric learning with structure-aware knowledge distillation (SA), significantly mitigating catastrophic forgetting.

\textbf{CCL} \cite{cui2023ralccl}: A continual LPR method employing asymmetric contrastive learning and incorporating a knowledge distillation loss based on feature distribution.

\textbf{MICL} \cite{liu2024micl}: A recent approach that introduces mutual information to LPR methods, preserving more domain-shared information through the proposed mutual information loss.

\begin{table}[t]
        \centering
        \caption{Datasets with detailed information}
        \label{tab:datasets}
        \setlength{\tabcolsep}{3pt}
        \resizebox{\columnwidth}{!}{
            \begin{tabular}{c|p{1.2cm}<{\centering}ccccc}
            \hline
            \multirow{2}*{Dataset}&LiDAR&\multirow{2}*{Location}&Train&Test&Database&Query\\
            ~&Sensor&~&Size&Size&Date&Date\\
            \hline
            \multirow{2}*{Oxford \cite{maddern20171}}&SICK LMS-151&\multirow{2}*{Oxford}&\multirow{2}*{22k}&\multirow{2}*{3k}&\multirow{2}*{05/2014}   & \multirow{2}*{11/2015}\\
            \multirow{2}*{DCC \cite{kim2020mulran}}       & Ouster OS1-64   & \multirow{2}*{Dajeon}    & \multirow{2}*{5.5k}  & \multirow{2}*{15k} & \multirow{2}*{08/2019}   & \multirow{2}*{09/2019}\\
            \multirow{2}*{Riverside \cite{kim2020mulran}}  & Ouster OS1-64   & \multirow{2}*{Sejong}    & \multirow{2}*{5.5k}  & \multirow{2}*{18.6k} & \multirow{2}*{08/2019}   & \multirow{2}*{08/2019}\\
            \multirow{2}*{In-house \cite{uy2018pointnetvlad}}  & Velodyne HDL-64 & \multirow{2}*{Singapore} & \multirow{2}*{6.7k}  & \multirow{2}*{1.8k} & \multirow{2}*{10/2017}   & \multirow{2}*{10/2017}\\
            \multirow{2}*{KITTI \cite{geiger2013vision}}     & Velodyne HDL-64 & \multirow{2}*{Karlsruhe} & \multirow{2}*{-}     & \multirow{2}*{14.4k} & \multirow{2}*{10/2011}   & \multirow{2}*{10/2011}\\
            \hline
            \end{tabular}
            }
            \vspace{-0.5cm}
    \end{table}

\subsection{Implementation Details}
Consistent with the training methods in InCloud \cite{knights2022irosincloud} and CCL \cite{cui2023ralccl}, we use the metric learning strategy only during the initial environment (Oxford) training, and employ the continual learning strategy in subsequent environments. 
Similar to CCL's \cite{cui2023ralccl} implementation of LoGG3D-Net \cite{vidanapathirana2022logg3d}, we replace the second-order pooling with the max pooling module and omit point-based local consistency losses, as they do not perform well on non-adjacent datasets.
We adopt a random sampling memory construction strategy consistent with InCloud \cite{knights2022irosincloud}.

In our experiments, we adopt the batch expansion mechanism proposed in MinkLoc3D \cite{komorowski2021minkloc3d}. 
The initial batch size is set to 16, and the final batch size is set to 256, with a batch expansion rate of 1.4. 
In the following continual steps, we train each backbone model for 60 epochs. 
The learning rate is initialized at $1e^{-4}$ and is reduced by a factor of 0.1 after the 30th epoch.
We use Adam as the optimizer with a weight decay of $1e^{-3}$ and set the memory buffer size to 256 during training. 
All experiments are conducted on the same machine with a single Nvidia GeForce RTX 3090 GPU to ensure a fair comparison.

\subsection{Evaluation Metric}
The evaluation metrics for each task include the commonly used place recognition metric, mean Recall@1, and the forgetting score. 
The forgetting score, introduced by InCloud \cite{knights2022irosincloud} and CCL \cite{cui2023ralccl}, measures how much the model has forgotten of what it learned in the past.
It can be defined as follows:
\begin{equation}
        F =  \frac{1}{T-1}\sum_{t = 1}^{T-1}( \max\limits_{l\in 1\dots t}\{R_{l,t}\} -R_{T,t}),
\end{equation}
where $T$ is the total number of training environments and $t$ is the current training task. $R_{l,t}$ represents the Recall@1 of the test set for task $t$ after training step $l$. 
A higher mean Recall@1 indicates better place recognition performance. 
A lower forgetting score reflects less performance degradation in the old domains.
For the continual evaluations on the seen tasks, we report both the mean Recall@1 and the forgetting score. 
To evaluate the generalization ability of different methods, we report only the Recall@1 on the KITTI dataset.

\subsection{Performance on Overcoming Catastrophic Forgetting}
\begin{figure}[t]
        \centering
        \includegraphics[width=1.0\columnwidth]{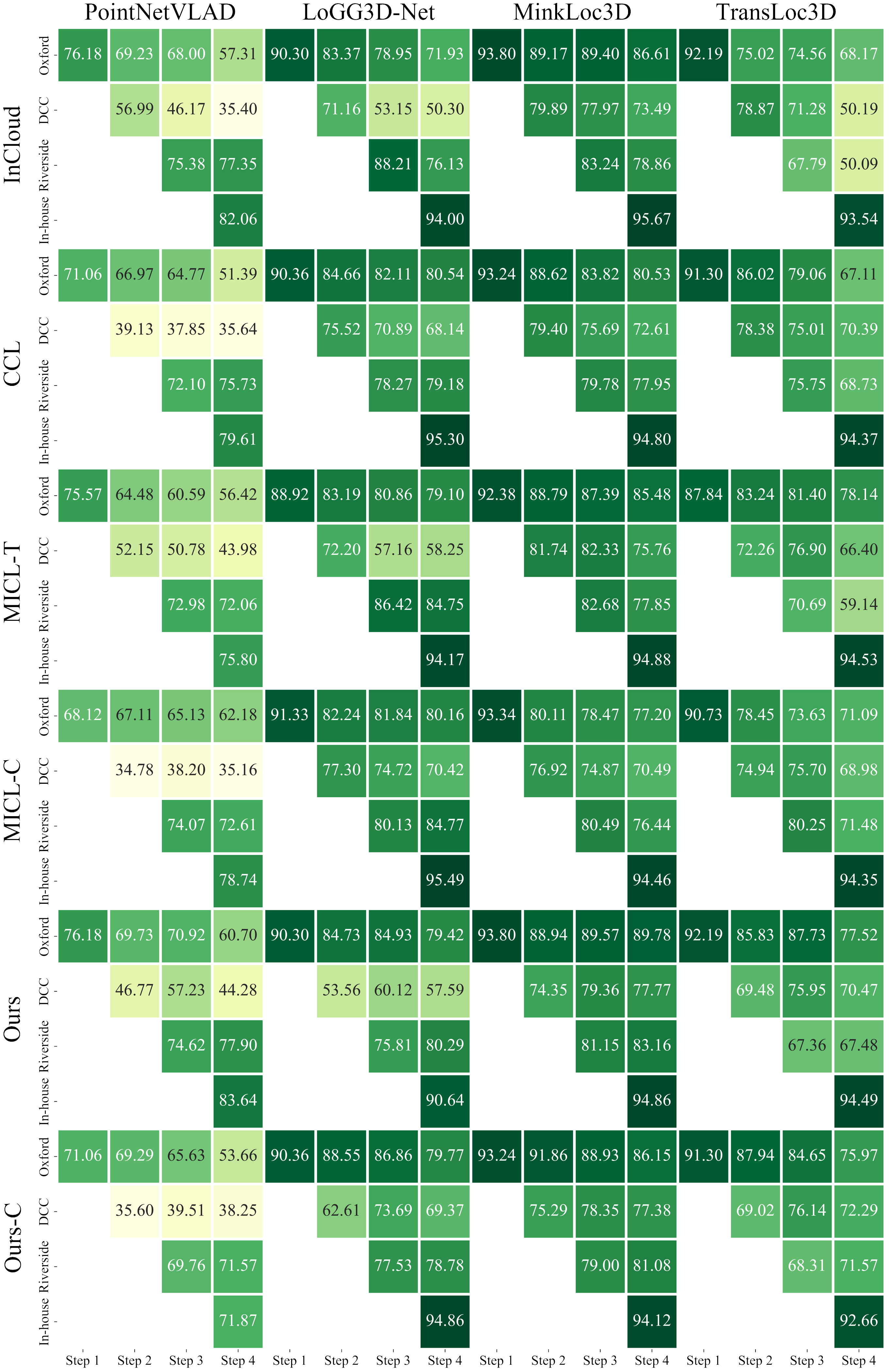}
        \caption{Detailed performance comparison in the $4$-$step$ protocol across PointNetVLAD, LoGG3D-Net, MinkLoc3D, and TransLoc3D with methods InCloud \cite{knights2022irosincloud}, CCL \cite{cui2023ralccl}, and MICL \cite{liu2024micl}. 
        The results are composed of Recall@1 matrices, where each row represents a continual learning method and each column represents a different LPR backbone. 
        In each Recall@1 matrix, steps 1 to 4 record the Recall@1 results of the environments that have been visited.
        A darker color means higher Recall@1.
        }\label{fig:detailed Recall@1}
        \vspace{-0.5cm}
\end{figure}

\begin{table*}[t]
        \centering
        \caption{Continual learning results on the first protocol}
        \label{tab:continual results}
        \belowrulesep=0pt
        \aboverulesep=0pt
        \renewcommand\arraystretch{1.1}
        \resizebox{\textwidth}{!}{
        \begin{threeparttable}
        \begin{tabular}{lcccccccccc}
        \hline
        \multirow{2}*{Methods}     &\multicolumn{2}{c}{PointNetVLAD}     &\multicolumn{2}{c}{LoGG3D-Net}     &\multicolumn{2}{c}{MinkLoc3D}     &\multicolumn{2}{c}{TransLoc3D}     &\multicolumn{2}{c}{Overall}    \\  \cmidrule(lr){2-3}\cmidrule(lr){4-5} \cmidrule(lr){6-7}\cmidrule(lr){8-9} \cmidrule(lr){10-11}
        ~&mR@1$\uparrow$&F$\downarrow$&mR@1$\uparrow$&F$\downarrow$&mR@1$\uparrow$&F$\downarrow$&mR@1$\uparrow$&F$\downarrow$&mR@1$\uparrow$&F$\downarrow$ \\ \hline
        $\textbf{Triplet Loss}$ &&&&&&&&&&\\
        Fine-Tuning    & 57.97& 21.28& 64.87& 30.28& 72.49& 22.40& 52.91& 39.79 & 62.06 & 28.44\\
        LwF \cite{li2017learning}           & 58.11& 20.49& 70.28& 21.81& 73.64& 19.82& 56.58& 34.8 & 64.65 & 24.23\\
        EWC \cite{kirkpatrick2017overcoming} & 58.21& 23.27& 69.47& 23.22& 78.14& 14.15& 60.16& 29.03 & 66.50 & 22.42\\
        InCloud \cite{knights2022irosincloud} & 63.03& 12.83& 73.09& 17.11& 83.66& 5.99& 65.49& 23.47& 71.32& 14.85\\
        MICL-T  \cite{liu2024micl} & 62.06 & 9.41 & $\textbf{79.07}$ & 8.48 & 83.49 & 6.10 & 74.55 & 10.58 & 74.79 & 8.64 \\

        Ours     & $\textbf{66.63}$ & $\textbf{8.38}$& 76.98& $\textbf{2.98}$& $\textbf{86.40}$ & $\textbf{1.20}$&$\textbf{77.49}$&$\textbf{6.68}$& $\textbf{76.88}$& $\textbf{4.92}$\\ \hline
        $\textbf{Contrastive Loss}$ &&&&&&&&&&\\
        CCL \cite{cui2023ralccl}    & 60.59& 6.51& 80.79& 5.43& 81.47& 7.11& 75.15& 13.07 & 74.5 & 8.03\\
        MICL-C  \cite{liu2024micl}    & $\textbf{62.17}$& $\textbf{3.48}$& $\textbf{82.71}$& $\textbf{4.47}$& 79.65& 8.87& 76.47& 11.71& 75.25& 7.13\\

        Ours-C & 58.84&5.61& 80.69& 4.56& $\textbf{84.68}$& $\textbf{1.99}$& $\textbf{78.12}$& $\textbf{5.31}$&$\textbf{75.58}$& $\textbf{4.37}$\\\hline
        \end{tabular}
        \begin{tablenotes}
                \item[*] The best results of the experiment are shown in \textbf{bold}.
        \end{tablenotes}
        \end{threeparttable}  
        }
        \vspace{-0.3cm}
\end{table*}

After training on a new environment, we report Recall@1 for both the current and previous environments.
\Reffig{fig:detailed Recall@1} shows the performance of InCloud \cite{knights2022irosincloud}, CCL \cite{cui2023ralccl}, and KDF on PointNetVLAD \cite{uy2018pointnetvlad}, LoGG3D-Net \cite{vidanapathirana2022logg3d}, MinkLoc3D \cite{komorowski2021minkloc3d}, and TransLoc3D \cite{xu2023transloc3d}, respectively.
To ensure a fair comparison of different methods, we have developed a contrastive loss variant of KDF (Ours-C). 
This variant uses the same asymmetric contrastive loss as CCL \cite{cui2023ralccl}, while also incorporating the ranking-aware knowledge distillation loss and the fusion module.

In \Reftab{tab:continual results}, we report the mean Recall@1 and forgetting score for methods based on triplet loss and contrastive loss, respectively.
MICL-T \cite{liu2024micl} and MICL-C \cite{liu2024micl} represent the results for triplet loss and contrastive loss, respectively.

Among the triplet loss-based methods, the mean Recall@1 of the classic continual learning methods LwF \cite{li2017learning} and EWC \cite{kirkpatrick2017overcoming} is only 2.59\% and 4.44\% higher than that of Fine-tuning, while their forgetting scores are 4.21\% and 6.02\% lower, respectively.
In contrast, InCloud \cite{knights2022irosincloud}, which utilizes structure-aware distillation, and KDF, which employs ranking-aware distillation, demonstrate significantly better performance in mitigating forgetting.
Overall, KDF outperforms other methods across all four backbone models. 
Compared to InCloud \cite{knights2022irosincloud}, KDF achieves a 5.56\% higher mean Recall@1 and a 9.93\% lower forgetting score. 
The detailed results are shown in the first and fifth rows of 4 by 4 blocks in \Reffig{fig:detailed Recall@1}, which illustrate the Recall@1 results of InCloud and our method in each environment under the $4$-$step$ protocol.
In each Recall@1 matrix, steps 1 to 4 indicate the Recall@1 results for the environments that have been visited.

Among methods based on contrastive loss, KDF surpasses CCL \cite{cui2023ralccl} in overall performance, achieving an average improvement of 0.92\% in mean Recall@1 and a 3.66\% lower forgetting score.
The detailed Recall@1 results for the four environments are presented in the second and sixth rows of the 4 by 4 blocks in \Reffig{fig:detailed Recall@1}.

Compared to the latest SOTA method MICL \cite{liu2024micl}, KDF outperforms the corresponding version of MICL overall.
For PointNetVLAD \cite{uy2018pointnetvlad} and LoGG3D-Net \cite{vidanapathirana2022logg3d}, our triplet loss version performs similarly to MICL-T \cite{liu2024micl}, whereas the contrastive loss version performs worse than MICL-C \cite{liu2024micl}.
This difference might be attributed to MICL's approach of maximizing mutual information between the current model and all previous models, whereas KDF only distills knowledge between two models (the old model and the current model).
For MinkLoc3D \cite{komorowski2021minkloc3d} and TransLoc3D \cite{xu2023transloc3d}, our KDF performs better than MICL \cite{liu2024micl}.

\subsection{Performance on Generalization Performance}

\begin{table}[t]
        \centering
        \caption{Generalization results on the KITTI dataset}
        \label{tab:generalization results}
        \setlength{\tabcolsep}{3pt}
        \normalsize
        \resizebox{\columnwidth}{!}{
        \begin{threeparttable}
        \begin{tabular}{lccccc}
        \hline
        Method       & PointNetVLAD     & LoGG3D-Net       & MinkLoc3D      & TransLoc3D  & Overall\\ \hline
        $\textbf{Sequence 00}$  & & & & &\\
        InCloud \cite{knights2022irosincloud}     & 88.78            & 91.62            & 91.36          & 89.82  & 90.40\\
        CCL \cite{cui2023ralccl}         & 87.76          & $\textbf{93.42}$  & $\textbf{92.53}$          & 90.33          & 91.01        \\
        MICL-T \cite{liu2024micl}  & $\textbf{91.75}$ & 92.27  & $\textbf{92.53}$  & $\textbf{91.49}$& $\textbf{92.01}$\\
        MICL-C \cite{liu2024micl}  & 90.08          & 92.78 & 92.27  & 90.07  & 91.30     \\
        Ours         & 90.85 & 91.88            & 91.62 & $\textbf{91.49}$ & 91.46\\ 
        Ours-C               & 90.08          & 92.65  & 92.40          & 90.33          & 91.37 
        \\ \hline
        $\textbf{Sequence 02}$  & & & & &\\
        InCloud  \cite{knights2022irosincloud}            & 68.90        & 74.58  & 81.94     & 73.24      & 74.67        \\
        CCL  \cite{cui2023ralccl}                & 62.54          & $\textbf{79.60}$          & 78.93          & 72.24          & 73.33        \\
        MICL-T \cite{liu2024micl}     & 69.23       & 70.57        & 82.54          & 72.91  & 73.81\\
        MICL-C \cite{liu2024micl}         & 69.23 & 77.59  & 77.93          & 74.92          & 74.92  \\
        Ours                 & $\textbf{71.91}$        & 73.91  & $\textbf{82.61}$     & 72.24      & $\textbf{75.17}$ \\
        Ours-C               & 69.57          & 76.59          & 76.25          & $\textbf{75.59}$          & 74.50        
        \\ \hline
        $\textbf{Sequence 07}$  & & & & &\\
        InCloud \cite{knights2022irosincloud}             & 60.71        & 60.71  & 64.29     & 67.86      & 63.39        \\
        CCL    \cite{cui2023ralccl}              & 57.14          & $\textbf{71.43}$          & $\textbf{75.00}$          & 71.43          & 68.75        \\
        MICL-T \cite{liu2024micl}     & 60.71    & 67.86& $\textbf{75.00}$ & 71.43 & 68.75\\
        MICL-C \cite{liu2024micl}     & 57.14  & $\textbf{71.43}$  & $\textbf{75.00}$   & 71.43  & 68.75 \\
        Ours                 & 57.14        & $\textbf{71.43}$  & 67.86     & 71.43       & 66.97  \\
        Ours-C               & $\textbf{71.43}$          & $\textbf{71.43}$          & 71.43          & $\textbf{75.00}$          & $\textbf{72.32}$
        \\ \hline
        $\textbf{Sequence 08}$  & & & & &\\
        InCloud \cite{knights2022irosincloud}             & 0.31         & 10.69  & $\textbf{49.69}$     & 43.08      & 25.94        \\
        CCL   \cite{cui2023ralccl}               & 5.35          & 53.14          & 42.45          & 43.40          & 36.09        \\
        MICL-T \cite{liu2024micl}     & $\textbf{10.38}$  & 16.67 & 49.37  & 40.88 & 29.33\\
        MICL-C \cite{liu2024micl}     & 3.77 & 49.06  & 40.25  & 37.74  &32.71 \\
        Ours                 & 1.57         & 12.89  & 48.11     & 41.82      & 26.10  \\
        Ours-C               & 5.66    & $\textbf{58.17}$          & 43.40          & $\textbf{43.71}$           & $\textbf{37.74}$         
        \\ \hline
        \end{tabular}
        \begin{tablenotes}
                \item[*] The best results of the experiment are shown in \textbf{bold}.
        \end{tablenotes}
        \end{threeparttable}  
        }
        \vspace{-0.5cm}
\end{table}

This experiment evaluates the generalization performance of InCloud \cite{knights2022irosincloud}, CCL \cite{cui2023ralccl}, MICL \cite{liu2024micl}, and KDF. 
\Reftab{tab:generalization results} reports the Recall@1 on the unseen domain (KITTI dataset) under the second protocol.
The LPR tasks for sequences 00 and 02 are relatively simple, so different methods perform well.
Sequence 07 has the fewest loops, while sequence 08 is the most challenging due to its reverse route direction.
In sequences 07 and 08, methods based on contrastive loss generally perform better than those based on triplet loss.
Our method surpasses InCloud \cite{knights2022irosincloud} and MICL-T \cite{liu2024micl} in average performance, and our contrastive loss variant also outperforms CCL \cite{cui2023ralccl} and MICL-C \cite{liu2024micl}.
Overall, KDF achieves the best performance on the 02, 07, and 08 sequences, and only performs weaker than MICL-T \cite{liu2024micl} on the 00 sequence. 
The experimental results demonstrate that KDF enables smooth knowledge transfer and learns effective place recognition.
After training on four datasets, KDF maintains a more generalized place recognition capability on the unseen dataset.

\subsection{Effect of Different Divergences Losses}
\label{sec:experiment_divergences}
In this section, we explore the effect of different divergence metrics as distribution-based distillation losses. 
We select KL divergence \cite{kullback1951information} and JS divergence \cite{fuglede2004jensen} for the experiment under the $4$-$step$ protocol. 
In this experiment, we replace only the SKL divergence loss in KDF, keeping the other modules and losses unchanged.
\Reftab{tab:divergences effect} presents the results for mean Recall@1 and forgetting score with different divergence metrics. 
Compared to KL divergence and JS divergence, KDF reports an improvement in mean Recall@1 and a reduction in the forgetting score, respectively. 
The experimental results demonstrate that the SKL divergence loss captures differences between feature distributions more accurately.

\begin{table}[t]
        \centering
        \caption{The effect of different divergences in the proposed method.}
        \label{tab:divergences effect}
        \belowrulesep=0pt
        \aboverulesep=0pt
        \setlength{\tabcolsep}{2pt}
        \renewcommand\arraystretch{1.2}
        \resizebox{\columnwidth}{!}{
        \begin{threeparttable}
        \begin{tabular}{ccccccccccc}
        \hline
        \multirow{2}*{Type}&\multicolumn{2}{c}{PointNetVLAD} &\multicolumn{2}{c}{LoGG3D-Net} &\multicolumn{2}{c}{MinkLoc3D} &\multicolumn{2}{c}{TransLoc3D} &\multicolumn{2}{c}{Overall}    \\
        \cmidrule(lr){2-3}\cmidrule(lr){4-5} \cmidrule(lr){6-7}\cmidrule(lr){8-9}\cmidrule(lr){10-11}
        ~&mR@1$\uparrow$&F$\downarrow$&mR@1$\uparrow$&F$\downarrow$&mR@1$\uparrow$&F$\downarrow$&mR@1$\uparrow$&F$\downarrow$&mR@1$\uparrow$&F$\downarrow$\\ \hline
        KL     & 66.40& 9.03& 76.45& 3.48& 84.71& 3.34& $\textbf{77.77}$& $\textbf{6.41}$& 76.33& 5.57\\
        JS     & 66.62& 9.00& $\textbf{78.07}$& 4.72& 85.08& 2.48& 77.38& 7.39& 76.79& 5.90\\
        Ours              & $\textbf{66.63}$ & $\textbf{8.38}$& 76.98& $\textbf{2.98}$& $\textbf{86.40}$ & $\textbf{1.20}$& 77.49& 6.68&$\textbf{76.88}$ &$\textbf{4.81}$ \\ \hline
        \end{tabular}
        \begin{tablenotes}
        \item[*] The best results of the experiment are shown in bold.
        \end{tablenotes}
        \end{threeparttable}  
        }
\end{table}

\subsection{Ablation Study}
To better understand the contribution of each module and how these modules work together, we report the results with respect to mean Recall@1 and the forgetting score on the first protocol using the MinkLoc3D \cite{komorowski2021minkloc3d} model in \Reftab{tab:ablation study}.
The base setting uses only the $\mathcal{L}_{PR}$ loss as supervision.
This experiment demonstrates that each component of KDF contributes to the LPR performance, and KDF achieves the best performance when all losses and proposed modules are used.
\begin{table}[t]
        \centering
        \caption{Ablation Study}
        \label{tab:ablation study}
        \resizebox{\columnwidth}{!}{
        \begin{threeparttable}
        \begin{tabular}{cccc|cc}
        \hline
        Base & $\mathcal{L}_{RKD}$ & $\mathcal{L}_{DKD} $ & Knowledge Fusion & mR@1$\uparrow$       & F$\downarrow$               \\ \hline
        $\surd $  &              &                   &                 & 81.54            & 7.99            \\
        $\surd $  & $\surd $     &                   &                 & 83.00            & 6.92            \\
        $\surd $  &              & $\surd $          &                 & 83.83            & 5.76            \\
        $\surd $  &              &                   & $\surd $        & 83.43            & 4.64            \\
        $\surd $  & $\surd $     & $\surd $          &                 & 84.38            & 5.17            \\
        $\surd $  & $\surd $     &                   & $\surd $        & 84.37            & 3.51            \\
        $\surd $  &              & $\surd $          & $\surd $        & 85.65            & 2.05            \\
        $\surd $  & $\surd $     & $\surd $          & $\surd $        & $\textbf{86.40}$ & $\textbf{1.20}$ \\ \hline
        \end{tabular}
        \begin{tablenotes}
        \item[*] The best results of the experiment are shown in bold.
        \end{tablenotes}
        \end{threeparttable} 
        }
\end{table}

\section{CONCLUSION}
\label{sec:conclusion}
In this paper, we introduce a continual learning framework, KDF, specifically designed for LiDAR place recognition. 
The KDF framework consists of two key components: knowledge distillation and knowledge fusion.
In knowledge distillation, we utilize a ranking-aware loss function to encourage the network to preserve place recognition knowledge. 
Additionally, we introduce a knowledge fusion module to integrate knowledge from both new and old tasks to mitigate catastrophic forgetting.
Through extensive experiments, we demonstrate the superior performance of KDF in alleviating forgetting. 
KDF can be applied to various networks to address forgetting, and it also exhibits enhanced generalization in new environments.
In future work, we will continue to explore mechanisms to extract more useful knowledge from the model and aim to extend this approach to visual place recognition (VPR) and multimodal place recognition tasks.
\bibliographystyle{IEEEtran}
\bibliography{IEEEabrv, paper}

\end{document}